\documentclass[8pt,twocolumn]{article}
\usepackage{longtable}
\usepackage{algorithm}
\usepackage{algorithmic}
\usepackage{booktabs}
\usepackage{amsmath}
\usepackage{amssymb}
\usepackage{amsthm}
\usepackage{bm}
\usepackage{boxedminipage}
\usepackage{dsfont}
\usepackage{graphicx}
\usepackage{natbib}
\usepackage{savesym}
\savesymbol{AND}
\usepackage{framed}
\usepackage[dvipdfmx,setpagesize=false]{hyperref}

\theoremstyle{definition}
\newtheorem{theorem}{Theorem}
\theoremstyle{definition}
\newtheorem{lemma}{Lemma}[section]
\theoremstyle{definition}

\theoremstyle{definition}

\theoremstyle{definition}

\theoremstyle{definition}
\newtheorem{definition}{Definition}
\theoremstyle{definition}
\newtheorem{proposition}{Proposition}
\theoremstyle{definition}
\newtheorem{property}{Property}
\theoremstyle{definition}

\theoremstyle{definition}
\newtheorem{conjecture}{Conjecture}[section]
\theoremstyle{definition}
\newtheorem{assumption}{Assumption}[section]
\theoremstyle{definition}
\newtheorem{corollary}{Corollary}[section]
\theoremstyle{definition}
\newtheorem{exercise}{Exercise}[section]
\theoremstyle{definition}
\newtheorem{simulation}{Simulation}[section]

 {\endMakeFramed}

\newenvironment{greenleftbar}{%
  \def\FrameCommand{\textcolor{green}{\vrule width 0.5em} \hspace{10pt}}%
  \MakeFramed {\advance\hsize-\width \FrameRestore}}%
 {\endMakeFramed}

\newenvironment{lightgrayleftbar}{%
  \def\FrameCommand{\textcolor{lightgray}{\vrule width 0.7em} \hspace{10pt}}%
  \MakeFramed {\advance\hsize-\width \FrameRestore}}%
{\endMakeFramed}

\newenvironment{exercise-waku}
  {\begin{lightgrayleftbar}\begin{exercise}\footnotesize }
  {\footnotesize \end{exercise}\end{lightgrayleftbar}}
\newenvironment{simulation-waku}
  {\begin{greenleftbar}\begin{simulation}}
  {\end{simulation}\end{greenleftbar}}
\newenvironment{proposition-waku}
  {\begin{oframed}\begin{proposition}}
  {\end{proposition}\end{oframed}}
\newenvironment{definition-waku}
  {\begin{oframed}\begin{definition}}
  {\end{definition}\end{oframed}}
\newenvironment{lemma-waku}
  {\begin{oframed}\begin{lemma}}
  {\end{lemma}\end{oframed}}
\newenvironment{theorem-waku}
  {\begin{oframed}\begin{theorem}}
  {\end{theorem}\end{oframed}}
\newenvironment{property-waku}
  {\begin{oframed}\begin{property}}
  {\end{property}\end{oframed}}
\newenvironment{corollary-waku}
  {\begin{oframed}\begin{corollary}}
  {\end{corollary}\end{oframed}}
\newenvironment{conjecture-waku}
  {\begin{oframed}\begin{conjecture}}
  {\end{conjecture}\end{oframed}}
\newenvironment{assumption-waku}
  {\begin{oframed}\begin{assumption}}
  {\end{assumption}\end{oframed}}

\newcommand{\0}{{\bm{0}}}
\newcommand{\1}{{\bm{1}}}

\newcommand{\vc}{{\bm{c}}}

\newcommand{\ve}{{\bm{e}}}

\newcommand{\vs}{{\bm{s}}}

\newcommand{\vu}{{\bm{u}}}

\newcommand{\vv}{{\bm{v}}}
\newcommand{\vw}{{\bm{w}}}

\newcommand{\x}{{\bm{x}}}

\newcommand{\y}{{\bm{y}}}

\newcommand{\z}{{\bm{z}}}

\newcommand{\vA}{{\bm{A}}}

\newcommand{\cA}{{\mathcal{A}}}

\newcommand{\cB}{{\mathcal{B}}}
\newcommand{\vC}{{\bm{C}}}
\newcommand{\cC}{{\mathcal{C}}}

\newcommand{\cD}{{\mathcal{D}}}
\newcommand{\bE}{{\mathbb{E}}}

\newcommand{\cE}{{\mathcal{E}}}

\newcommand{\vI}{{\bm{I}}}
\newcommand{\cI}{{\mathcal{I}}}

\newcommand{\cL}{{\mathcal{L}}}

\newcommand{\bN}{{\mathbb{N}}}

\newcommand{\vO}{{\bm{O}}}

\newcommand{\bR}{{\mathbb{R}}}

\newcommand{\cS}{{\mathcal{S}}}

\newcommand{\vS}{{\bm{S}}}

\newcommand{\vW}{{\bm{W}}}

\newcommand{\cY}{{\mathcal{Y}}}
\newcommand{\X}{{\bm{X}}}

\newcommand{\Z}{{\bm{Z}}}

\newcommand{\valph}{{\bm{\alpha}}}
\newcommand{\vbeta}{{\bm{\beta}}}

\newcommand{\vDel}{{\bm{\Delta}}}

\newcommand{\argmax}{\mathop{\textrm{argmax}}\limits}
\newcommand{\argmin}{\mathop{\textrm{argmin}}\limits}

 {\end{tabular}\end{flushleft}}

\newenvironment{tsaligned}{\begin{equation}\begin{aligned}}{\end{aligned}\end{equation}}
\newcommand{\rloss}{r_{\text{loss}}}

\begin{document}

\twocolumn[{%
\begin{center}
{\Large
  Sign-Constrained Regularized Loss Minimization}
\\
\vspace{1cm}
       {\large
         Tsuyoshi Kato${}^{\dagger,*}$, 
         Misato Kobayashi${}^{\dagger}$, 
         Daisuke Sano${}^{\diamond}$}
\\
\vspace{1cm}
\begin{tabular}{lp{0.7\textwidth}}
${}^\dagger$ & 
Division of Electronics and Informatics, Faculty of Science and Technology, Gunma University, Tenjin-cho 1-5-1, Kiryu, Gunma 376-8515, Japan. 
\\
${}^\diamond$ &
Department of Civil and Environmental Engineering, Graduate School of Engineering, Tohoku University, Aoba 6--6--06, Aramaki, Aoba-ku, Sendai, Miyagi 980--8579, Japan. 
\\
\end{tabular}
\end{center}
}]
\sloppy

\begin{abstract}
  In practical analysis, domain knowledge about analysis target has often been accumulated, although, typically, such knowledge has been discarded in the statistical analysis stage, and the statistical tool has been applied as a black box. In this paper, we introduce sign constraints that are a handy and simple representation for non-experts in generic learning problems. We have developed two new optimization algorithms for the sign-constrained regularized loss minimization, called the sign-constrained Pegasos (SC-Pega) and the sign-constrained SDCA (SC-SDCA), by simply inserting the sign correction step into the original Pegasos and SDCA, respectively.  We present theoretical analyses that guarantee that insertion of the sign correction step does not degrade the convergence rate for both algorithms.  Two applications, where the sign-constrained learning is effective, are presented. The one is exploitation of prior information about correlation between explanatory variables and a target variable. The other is introduction of the sign-constrained to SVM-Pairwise method.  Experimental results demonstrate significant improvement of generalization performance by introducing sign constraints in both applications.  
\end{abstract}

\section{Introduction}
The problem of regularized loss minimization
(e.g. \citet{HasTibFri-book09a}) is often described as 
\begin{tsaligned}
  \label{eq:prob-rlm-uncon}
  \text{min }\quad&
  P(\vw)\qquad\text{wrt }\quad\vw\in\bR^{d},
  \\
  \text{where }\quad&
  P(\vw) :=
  \frac{\lambda}{2}\lVert\vw\rVert^{2}
  +
  \frac{1}{n}\Phi(\X^\top\vw), 
  \\
  &
  \X := \left[\x_{1},\dots,\x_{n}\right]\in\bR^{d\times n},  
\end{tsaligned}
aiming to obtain a linear predictor
$\left<\vw,\x\right>$ for an unknown input $\x\in\bR^{d}$.  
Therein, $\Phi :\bR^{n}\to\bR$ is a loss function
which is the sum of convex losses for $n$ examples:
$\Phi(\z) := \sum_{i=1}^{n}\phi_{i}(z_{i})$
for $\z := \left[z_{1},\dots,z_{n}\right]^\top\in\bR^{n}$.
This problem covers a large class of machine learning algorithms including support vector machine, logistic regression, support vector regression, and ridge regression. 

In this study, we pose \emph{sign constraints}
\citep{Lawson1995solving}
to the entries in the model parameter $\vw\in\bR^{d}$
in the unconstrained minimization
problem~\eqref{eq:prob-rlm-uncon}.
We divide the index set of $d$ entries
into three exclusive subsets,
$\cI_{+}$, $\cI_{0}$, and $\cI_{-}$, as
$\{1,\dots,d\}
 =
 \cI_{+}\cup\cI_{0}\cup\cI_{-}$
and impose on the entries in
$\cI_{+}$ and $\cI_{-}$, 
\begin{tsaligned}\label{eq:sgncon}
  &\text{for }h\in\cI_{+},\quad w_{h}\ge 0,
  &&
  \text{for }h'\in\cI_{-},\quad w_{h'}\le 0. 
\end{tsaligned}

Sign constraints can introduce prior knowledge directly to learning machines. For example, let us consider a binary classification task. In case that $h$-th explanatory variable $x_{h}$ is positively correlated to a binary class label $y\in\{\pm 1\}$, then a positive weight coefficient $w_{h}$ is expected to achieve a better generalization performance than a negative coefficient, because without sign constraints, the entry $w_{h}$ in the optimal solution might be negative due to small sample problem.  On the other hand, in case that $x_{h}$ is negatively correlated to the class label, a negative weight coefficient $w_{h}$ would yield better prediction.  If sign constraints were explicitly imposed, then inadequate signs of coefficients could be avoided. 


The strategy of sign constraints for generic learning problems has rarely been discussed so far, although there are extensive reports for non-negative least square regression supported by many successful applications including sound source localization: \citep{YuanqingLin2004-icassp}, tomographic imaging \citep{JunMa2013-algo}, spectral analysis \citep{QiangZhang07-asrc}, hyperspectral image super-resolution \citep{DonFuShi16}, microbial community pattern detection \citep{CaiGuKen17}, face recognition \citep{YangfengJi2009-icmla,HeZheHu13}, and non-negative image restoration \citep{Henrot2013-icassp,Landi2012-na,YanfeiWang2007-ipse,Shashua2005-icml}. 
In most of them, non-negative least square regression is used as an important ingredient of bigger methods such as non-negative matrix factorization \citep{lee2001algorithms,WanTiaYu17,Kimura2016column,Fvotte2011algo,Ding2006ortho}.

Several efficient algorithms for the non-negative least square regression have been developed. The active set method by \citet{Lawson1995solving} has been widely used in many years, and several work \citep{DongminKim2010-siamjsc,DongminKim2007-siam,Bierlaire1991-laa,Portugal1994comparison,More91-siamjo,ChihJenLin1999-siamjo,Morigi2007-joam} have accelerated optimization by combining the active set method with the projected gradient approach.  Interior point methods~\citep{Bellavia2006,Heinkenschloss99-mp,Kanzow06-coa} have been proposed as an alternative algorithm for non-negative least square regression. However, all of them cannot be applied to generic regularized loss minimization problems. 

In this paper, we present two algorithms for the sign-constrained regularized loss minimization problem with generic loss functions.
A surge of algorithms for unconstrained regularized empirical loss minimization have been developed such as
  SAG \citep{Roux12a-sag,Schmidt2016-sag}, 
  SVRG \citep{Johnson13a-svrg}, 
  Prox-SVRG \citep{LinXiao2014-siamjo}, 
  SAGA \citep{Defazio2014-nips}, 
  Kaczmarz \citep{Needell2015}, 
  EMGD \citep{LijunZhang2013-nips}, and
  Finito \citep{defazio2014finito}. This study focuses on two popular algorithms, Pegasos~\citep{Shalev-Shwartz11-pegasos} and SDCA~\citep{Shalev-Shwartz2013a-SDCA}. A prominent characteristic of the two algorithms is unnecessity to choose a step size. Some of the other optimization algorithms guarantee convergence to the optimum under the assumption of a small step size, although the step size is often too small to be used. Meanwhile, the theorem of Pegasos has been developed with a step size $\eta_{t}=1/(\lambda t)$ which is large enough to be adopted actually. SDCA needs no step size. Two new algorithms developed in this study for the sign-constrained problems are simple modifications of Pegasos and SDCA.

The contributions of this study are summarized as follows.
\begin{itemize}
\item
  Sign constraints are introduced to generic regularized loss minimization problems. 
\item
  Two optimization algorithms for the sign-constrained regularized loss minimization, called \emph{SC-Pega} and \emph{SC-SDCA}, were developed by simply inserting the \emph{sign correction step}, introduced in Section~\ref{s:scpega}, to the original Pegasos and SDCA. 
\item
  Our theoretical analysis ensures that both SC-Pega and SC-SDCA do not degrade the convergence rates of the original algorithms.
\item
  Two attractive applications, where the sign-constrained learning is effective, are presented. The one is exploitation of prior information about correlation between explanatory variables and a target variable. The other is introduction of the sign-constrained to SVM-Pairwise method~\citep{LiaNob03-jcb}.
\item Experimental results demonstrate significant improvement of generalization performance by introducing sign constraints in both two applications.  
\end{itemize}
\section{Problem Setting}
The feasible region can be expressed simply as
\begin{tsaligned}\label{eq:fearegion}
  \cS :=
  \left\{ \vw\in\bR^{d}\,\middle|\,
  \vc\odot\vw\ge \0_{d}\right\} 
\end{tsaligned}
where 
$\vc = \left[ c_{1},\dots,c_{d} \right]^\top\in\{0,\pm 1\}^{d}$, 
each entry is given by
\begin{tsaligned}\label{eq:scvec-c-def}
  c_{h} :=
  \begin{cases}
    +1 \qquad&\text{for }h\in\cI_{+},
    \\
    0 \qquad&\text{for }h\in\cI_{0},
    \\
    -1 \qquad&\text{for }h\in\cI_{-}. 
  \end{cases}
\end{tsaligned}
Using $\cS$, the optimization problem
discussed in this paper can be expressed as
\begin{tsaligned}\label{eq:prob-rlm-signcon}
  \text{min }\quad&
  P(\vw)\qquad\text{wrt }\quad\vw\in\cS. 
\end{tsaligned}
\begin{assumption-waku}\label{assum:four-for-rlm-signcon}
  Throughout this paper, the following assumptions are used:
\begin{align*}
  \text{(a) } & \text{$\Phi(\cdot)$ is a convex function.}
  &
  \text{(b) } & \text{$\frac{1}{n}\Phi(\0)\le \rloss$.}
  \\
  \text{(c) } & \text{$\forall \vs\in\bR^{n}$, $\Phi(\vs)\ge 0$.}
  &
  \text{(d) } & \text{$\forall i$, $\lVert\x_{i}\rVert \le R$. }
\end{align*}
\end{assumption-waku}
\begin{table*}
  \caption{Loss functions and their properties. 
    Suppose $0\le \gamma\le 1$.
    Let $\y := \left[ y_{1},\dots,y_{n}\right]^\top$.
    \label{tab:loss}}
  \begin{center}
    \begin{tabular}{|c|l|c|c|c|}
      \hline
      Name & Definition & Label & Type & $\rloss$
      \\
      \hline
      Classical hinge loss &
      $\phi_{i}(s) := \max(0,1-y_{i}s)$ &
      $y_{i}\in\{\pm 1\}$ &
      $1$-Lipschitz &
      $1$
      \\
      \hline
      Smoothed hinge loss &
      \(\phi_{i}(s) :=
      \begin{cases}
        1 - y_{i}s -0.5\gamma \,\, &\text{ if } y_{i}s\in(-\infty,1-\gamma],
        \\
        (1-y_{i}s)^{2}/(2\gamma)  \,\, &\text{ if } y_{i}s\in(1-\gamma,1),        
        \\
        0 &\text{ if } y_{i}s\in [1,+\infty). 
      \end{cases}\)
      &
      $y_{i}\in\{\pm 1\}$ &
      $(1/\gamma)$-smooth &
      $1-\frac{\gamma}{2}$
      \\
      \hline
      Logistic loss &
      $\phi_{i}(s) := \log( 1 + \exp(-y_{i}s))$ &
      $y_{i}\in\{\pm 1\}$ &
      $0.25$-smooth &
      $\log(2)$
      \\
      \hline
      Square error loss
      &
      $\phi_{i}(s) := 0.5( s - y_{i} )^{2}$ & 
      $y_{i}\in\bR$ &
      $1$-smooth &
      $\lVert\y\rVert^{2}/(2n)$
      \\
      \hline
      Absolute error loss
      &
      $\phi_{i}(s) := | s - y_{i} |$ & 
      $y_{i}\in\bR$ &
      $1$-Lipschitz &
      $\lVert\y\rVert_{1}/n$
      \\
      \hline
    \end{tabular}
  \end{center}
\end{table*}
Most of widely used loss functions satisfy
the above assumptions.  Several examples of such
loss functions are described in Table~\ref{tab:loss}. 
If the hinge loss is chosen, the learning machine is a well-known instance called the
support vector machine. 
If the square error loss is chosen, the learning machine is called the
ridge regression. 
We denote the optimal solution to the constraint
problem by
$\vw_{\star}:=\argmin_{\vw\in\cS}P(\vw). $
We assume two types of loss functions:
\emph{$L$-Lipschitz continuous} function and
\emph{$(1/\gamma)$-smooth} function. 
Function $\phi_{i}:\bR\to\bR$ is said to be an $L$-Lipschitz continuous funciton
if 
\begin{tsaligned}\label{eq:Lipschitz-def}
  \forall s,\forall \delta \in\bR,\qquad
  |\phi_{i}(s+\delta)-\phi_{i}(s)| \le L |\delta|. 
\end{tsaligned}
Such functions are often said shortly to be \emph{$L$-Lipschitz}
in this paper.  
Function $\phi_{i}:\bR\to\bR$ is a $(1/\gamma)$-smooth
function if its derivative function is $L$-Lipschitz. 
%
%
For an index subset $\cA\subseteq\{1,\dots,n\}$
and a vector $\vv\in\bR^{n}$, 
let $\vv_{\cA}$ be the subvector of $\vv$ containing
entries corresponding to $\cA$.  Let $\X_{\cA}$
be a sub-matrix in $\X$ containing columns
corresponding to $\cA$.
Let $\Phi(\cdot\,;\,\cA):\bR^{|\cA|}\to\bR$ be defined as
\begin{tsaligned}
  \Phi(\vs_{\cA}\,;\,\cA)
  :=
  \sum_{i\in\cA}\phi_{i}(s_{i}). 
\end{tsaligned}

\section{Sign-Constrained Pegasos}\label{s:scpega}
In the original Pegasos algorithm~\citep{Shalev-Shwartz11-pegasos},
$\phi_{i}$ is assumed to be the classical hinge loss function
(See Table~\ref{tab:loss} for the definition).
Each iterate consists of three steps:
the \emph{mini-batch selection step},
the \emph{gradient step}, and
the \emph{projection-onto-ball step}.  
Mini-batch selection step chooses
a subset $\cA_{t}\subseteq\{1,\dots,n\}$
from $n$ examples at random.
The cardinality of the subset is predefined
as $|\cA_{t}| = k$. 
Gradient step computes the gradient of 
\begin{tsaligned}
  P_{t}(\vw) :=
  \frac{\lambda}{2}\lVert\vw\rVert^{2}
  + \frac{1}{k}\Phi(\X_{\cA_{t}}^\top\vw\,;\,\cA_{t}).   
\end{tsaligned}
which approximates the objective function $P(\vw)$. 
The current solution $\vw_{t}$ is moved toward
the opposite gradient direction as 
\begin{tsaligned}
  \vw_{t+1/2} &:=
  \vw_{t} - \frac{1}{\lambda t}\nabla P_{t}(\vw_{t})
  \\
  &=
  \frac{t-1}{t}
  \vw -
  \frac{1}{k\lambda t}
  \X_{\cA_{t}}
  \nabla\Phi(\X_{\cA_{t}}^\top\vw_{t}\,;\,\cA_{t}). 
\end{tsaligned}
At the projection-onto-ball step,
the norm of the solution is shortened to be
$\frac{1}{\sqrt{\lambda}\lVert\vw_{t+1/2}\rVert}$
if the norm is over
$\frac{1}{\sqrt{\lambda}\lVert\vw_{t+1/2}\rVert}$:  
\begin{tsaligned}
  \vw_{t+1} :=
  \min
  \left(
  1,
  \frac{1}{\sqrt{\lambda}\lVert\vw_{t+1/2}\rVert}
  \right)
  \vw_{t+1/2}. 
\end{tsaligned}
The projection-onto-ball step
plays an important role in getting
a smaller upper-bound
of the norm of the gradient of the
regularization term in the objective,
which eventually reduces the number of iterates
to attain an $\epsilon$-approximate solution
(i.e. $P(\tilde{\vw})-P(\vw_{\star})\le\epsilon$). 

In the algorithm developed in this study,
we simply inserts between those two steps,
a new step that corrects the sign of each
entry in the current solution~$\vw$ as
\begin{tsaligned}
  w_{h}\leftarrow
  \begin{cases}
    \max(0,w_{h})\quad&\text{for }h\in\cI_{+},
    \\
    \min(0,w_{h})\quad&\text{for }h\in\cI_{-},
    \\
    w_{h}\quad&\text{for }h\in\cI_{0},  
  \end{cases}
\end{tsaligned}
which can be rewritten equivalently as
$\vw \leftarrow \vw + \vc\odot(-\vc\odot\vw)_{+}$
where the operator $(\cdot)_{+}$ is defined as
$\forall \x\in\bR^{d}$,
$(\x)_{+}:=\max(\0,\x)$. 

\begin{algorithm}[t!]
\caption{
Generic Sign-Constrained Pegasos 
\label{alg:srpega-mbtch}}
\begin{algorithmic}[1]
  \REQUIRE
  Data matrix $\X\in\bR^{d\times n}$,
  loss function $\Phi:\bR^{n}\to\bR$,
  regularization parameter $\lambda\in\bR$,
  sign constraint parameter $\vc\in\{\pm 1,0\}^{d}$, and
  mini-batch size $k$. 
  \STATE \textbf{begin}
  \STATE $\vw_{1}:=\0_{d}$;  \{Initialization\}
  \FOR{$t:=1,\dots,T$}
  \STATE Choose $\cA_{t}\subseteq\{1,\dots,n\}$ uniformly
  at random such that $|A_{t}|=k$.
  \STATE
  $\vw_{t+1/3}
  :=
  \frac{t-1}{t}\vw_{t} -
  \frac{1}{\lambda t}\X_{\cA_{t}}\nabla\Phi(\X_{\cA_{t}}^\top\vw_{t-1};\cA_{t})$; 
  \STATE
  $\vw_{t+2/3}
  :=
  \vw_{t+1/3}
  + \vc\odot(-\vc\odot\vw_{t+1/3})_{+}$; 
  \STATE
  $\vw_{t+1}
  :=
  \min\left(
  1, \sqrt{\rloss\lambda^{-1}}\lVert\vw_{t+2/3}\rVert^{-1}
  \right)
  \vw_{t+2/3}$; 
  \ENDFOR
  \STATE \textbf{return }  $\tilde{\vw}:=\sum_{t=1}^{T}\vw_{t}/T$; 
  \STATE \textbf{end.}
\end{algorithmic}
\end{algorithm}
The algorithm can be summarized as
Algorithm~\ref{alg:srpega-mbtch}.
Here, the loss function is not limited to
the classical hinge loss. 
In the projection-onto-ball step, the solution
is projected onto 
$\sqrt{\rloss\lambda^{-1}}$-ball
instead of
$(1/\sqrt{\lambda})$-ball
to handle more general settings.
Recall that $\rloss=1$ if 
$\phi_{i}$ is the hinge loss
employed in the original Pegasos.
It can be shown that the 
objective gap is bounded as follows. 
\begin{theorem-waku}\label{thm:srpega-bound}
  Consider Algorithm~\ref{alg:srpega-mbtch}. 
  If $\phi_{i}$ are $L$-Lipschitz continuous,  
  it holds that
  \begin{tsaligned}
    \bE\left[P(\tilde{\vw})\right] - P(\vw_{\star}) \le
    \left( \sqrt{\rloss\lambda} + L R \right)^{2}
    \frac{1 + \log(T)}{\lambda T}. 
  \end{tsaligned}
\end{theorem-waku}
See Subsection~\ref{ss:proof-thm:srpega-bound} for proof
of Theorem~\ref{thm:srpega-bound}. 
This bound is exactly same as the original Pegasos, 
yet Algorithm~\ref{alg:srpega-mbtch} contains
the sign correction step. 

\section{Sign-Constrained SDCA}
The original SDCA is a framework for the unconstrained problems~\eqref{eq:prob-rlm-uncon}.
In SDCA, a dual problem is solved instead of the primal problem. Namely, the dual objective is maximized in a iterative fashion with respect to the dual variables $\valph:=\left[\alpha_{1},\dots,\alpha_{n}\right]^\top\in\bR^{n}$.
The problem dual to the unconstrained problem~\eqref{eq:prob-rlm-uncon} is given by
\begin{tsaligned}
  \label{eq:dual-rlm-uncon}
  \text{min }\quad&
  D(\valph)\qquad\text{wrt }\quad\valph\in\bR^{n},
\end{tsaligned}
where
\begin{tsaligned}
  &D(\valph) :=
  -
  \frac{\lambda}{2}\left\lVert\frac{1}{\lambda n}\X\valph\right\rVert^{2}
  -
  \frac{1}{n}\Phi^{*}(-\valph). 
\end{tsaligned}
%
To find the maximizer of $D(\valph)$, a single example $i$ is chosen randomly at each iterate $t$, and a single dual variable $\alpha_{i}$ is optimized with the other $(n-1)$ variables
$\alpha_{1},\dots,\alpha_{i-1}$, $\alpha_{i+1},\dots,\alpha_{n}$ frozen. If we denote by $\valph^{(t-1)}\in\bR^{n}$ the value of the dual vector at the previous iterate $(t-1)$, the dual vector is updated as
$\valph^{(t)}:=\valph^{(t-1)}+\Delta\alpha\ve_{i}$ where $\Delta\alpha\in\bR$ is determined so that $\Delta \alpha \in \argmax_{\Delta \alpha\in\bR}D_{t}(\Delta \alpha\,;\,\vw^{(t-1)})$ where $\vw^{(t-1)}=\frac{1}{\lambda n}\X\valph^{(t-1)}$ and
\begin{multline}
  D_{t}(\Delta \alpha\,;\,\vw)
  :=
  -\frac{\lambda}{2}\left\lVert\vw
  + \frac{\Delta \alpha}{\lambda n}\x_{i}\right\rVert^{2}
  -\frac{1}{n}\phi^{*}_{i}(-\alpha^{(t-1)}_{i}-\Delta \alpha). 
\end{multline}
In case of the hinge loss, the maximizer of $D_{t}(\cdot\,;\,\vw^{(t-1)})$ can be found within $O(d)$ computation. The primal variable $\vw^{(t)}$ can also be maintained within $O(d)$ computation by $\vw^{(t)}:=\vw^{(t-1)} + \frac{\Delta\alpha}{\lambda n}\x_{i}$.

Now let us move on the sign-constrained problem. In addition to Algorithm~\ref{alg:srpega-mbtch} that is derived from Pegasos, we present another algorithm based on SDCA for solving the minimizer of $P(\vw)$ subject to the sign constraint $\vc\odot\vw\ge\0_{d}$. Like Algorithm~\ref{alg:srpega-mbtch} that has been designed by inserting the sign correction step into the original Pegasos iterate, the new algorithm has been developed by simply adding the sign correction step in each SDCA iterate. The resultant algorithm is described in Algorithm~\ref{alg:srsdca}. 
\begin{algorithm}[t!]
\caption{
Generic Sign-Constrained SDCA. 
\label{alg:srsdca}}
\begin{algorithmic}[1]
  \REQUIRE
  Data matrix $\X\in\bR^{d\times n}$,
  loss function $\Phi:\bR^{n}\to\bR$,
  regularization parameter $\lambda\in\bR$, and
  sign constraint parameter $\vc\in\{\pm 1,0\}^{d}$. 
  \STATE \textbf{begin}
  \STATE $\valph^{(0)}:=\0_{n}$; $\bar{\vw}^{(0)}:=\0_{d}$; $\vw^{(0)}:=\0_{d}$;  \{Initialization\}
  \FOR{$t:=1,\dots,T$}
  \STATE
  $\Delta \alpha \in \argmax_{\Delta \alpha\in\bR}D_{t}(\Delta \alpha\,;\,\vw^{(t-1)})$; 
  \STATE
  $\bar{\vw}^{(t)}:=\bar{\vw}^{(t-1)} + \frac{\Delta\alpha}{\lambda n}\x_{i}$; 
  \STATE
  $\vw^{(t)} :=
  \bar{\vw}^{(t)}
  + \vc\odot(-\vc\odot\bar{\vw}^{(t)})_{+}$; 
  \ENDFOR
  \STATE
  \qquad \textbf{return } $\tilde{\vw}:=\frac{1}{T-T_{0}}\sum_{t=T_{0}+1}^{T}\vw^{(t-1)}$;  
  \STATE \textbf{end.}
\end{algorithmic}
\end{algorithm}

For some loss functions, maximization at step 5 in Algorithm~\ref{alg:srsdca} cannot be given in a closed form.
Alternatively, step~4 can be replaced to
\begin{tsaligned}\label{eq:sdca-prox-update}
  {\footnotesize 4: }&\quad
  \Delta\alpha := sq, \qquad \text{where}
  \\
  &s :=
  \text{Clip}_{[0,s^{-1}_{\text{lb}}]}
  \left(
  \frac{1}{2} +
  \frac{z^{(t)}\alpha_{i} + \phi_{i}^{*}(-\alpha_{i}) + \phi_{i}(z^{(t)})}{\gamma q^{2}}
  \right)s_{\text{lb}}.
\end{tsaligned}
Therein, we have defined 
$s_{\text{lb}}:=\lambda n \gamma / (\lambda n \gamma+R^{2})$,
$z^{(t)}:=\left<\vw^{(t-1)},\x_{i}\right>$,  
$q^{(t)}:=-\nabla\phi_{i}(z^{(t)})-\alpha_{i}^{(t-1)}$, and
$\text{Clip}_{[a,b]}(x) := \max(a,\min(b,x))$. 
See Subsection~\ref{ss:deriv-eq:sdca-prox-update} for derivation
of \eqref{eq:sdca-prox-update}. 

We have found the following theorem that
states the required number of iterates
guaranteeing 
the expected primal objective gap below a threshold $\epsilon$
under the sign constraints. 
\begin{theorem-waku}\label{thm:converg-srsdca}
  Consider Algorithm~\ref{alg:srsdca}. 
  In case that $\phi_{i}$ are $L$-Lipschitz continuous
  (i.e. \eqref{eq:Lipschitz-def}), 
  it holds that $\bE[P(\tilde{\vw})]-P(\vw_{\star})\le\epsilon$ if
  $T$ and $T_{0}$ are specified so that 
  \begin{tsaligned}
    T_{0}\ge \frac{4G}{\lambda \epsilon}
    + \max\left\{ 0, \left\lceil n\log\frac{2\lambda n \rloss}{G}\right\rceil\right\}
  \end{tsaligned}
  and
  \begin{tsaligned}
    T \ge T_{0}
    + \max\left\{ n, \frac{G}{\lambda \epsilon}\right\}
  \end{tsaligned}
  where $G:=4R^{2}L^{2}$.  If $\phi_{i}$ are hinge loss functions,
  then $G:=R^{2}L^{2}$. 
  In case that $\phi_{i}$ are $(1/\gamma)$-smooth,
  $\bE[P(\tilde{\vw})]-P(\vw_{\star})\le\epsilon$ is established if
  \begin{tsaligned}
    T > T_{0} \ge
    \left( n + \frac{R^{2}}{\lambda \gamma} \right)
    \log
    \left(
    \left( n + \frac{R^{2}}{\lambda \gamma} \right)
    \frac{\rloss}{(T-T_{0})\epsilon}
    \right).
  \end{tsaligned}
\end{theorem-waku}
See Subsections~\ref{ss:proof-thm:converg-srsdca}
for proof of 
Theorem~\ref{thm:converg-srsdca}.
Theorem~\ref{thm:converg-srsdca} suggests that the convergence rate of Algorithm~\ref{alg:srsdca} is not deteriorated compared to the original SDCA in both cases of $L$-Lipschitz and smooth losses, despite insertion of the sign correction step. 


\section{Multiclass Classification}
In this section, we extend our algorithms to the multi-class classification setting of $m$ classes. Here, the model parameter is a $\vW\in\bR^{d\times m}$ instead of a vector $\vw\in\bR^{d}$. The loss function for each example $\x_{i}\in\bR^{d}$ is of an $m$-dimensional vector. Here, the prediction is supposed to be done by taking the class with the maximal score among $s_{1}:=\left<\vw_{1},\x\right>, \dots$, and $s_{m}:=\left<\vw_{m},\x\right>$. Here, without loss of generality, the set of the class labels are given by $\cY:=\{1,\dots,m\}$.
Several loss functions $\phi^{\text{m}}_{i}:\bR^{m}\to\bR$
are used for multiclass classification as follows.
\begin{itemize}
\item Soft-max loss: 
\begin{tsaligned}
  \phi^{\text{m}}_{i}(\vs)
  :=
  \log\left( \sum_{y\in\cY} \exp\left( s_{y} -s_{y_{i}} \right) \right) 
\end{tsaligned}
Therein, $y_{i}$ is the true class label of $i$-th example.
\item Max-hinge loss; 
\begin{tsaligned}
  \phi^{\text{m}}_{i}(\vs)
  :=
  \max_{y\in\cY}
  \left( s_{y} -s_{y_{i}} + \delta_{y,y_{i}} \right). 
\end{tsaligned}
\item Top-$k$ hinge loss~\citep{lapin-nips2015}: 
\begin{tsaligned}
  \phi^{\text{m}}_{i}(\vs)
  :=
  \frac{1}{k}
  \sum_{j=1}^{k}
  \left( (\vI-\1\ve_{y_{i}}^\top)\vs + \1 - \ve_{y_{i}} \right)_{[j]}.
\end{tsaligned}
Therein, $x_{[j]}$ denotes the $j$-th largest value in a vector~$\x\in\bR^{m}$. 
\end{itemize}
The objective function for learning $\vW\in\bR^{d\times m}$
is defined as
\begin{tsaligned}
  P^{\text{m}}(\vW) :=
  \frac{\lambda}{2}\lVert\vW\rVert_{\text{F}}^{2}
  +
  \frac{1}{n}\sum_{i=1}^{n}\phi^{\text{m}}_{i}(\vW^\top\x_{i}). 
\end{tsaligned}
The learning problem discussed is
minimization of $P^{\text{m}}(\vW)$
with respect to $\vW$ 
subject to sign constraints
\begin{tsaligned}
  \forall(h,j)\in\cE_{+},\quad
  &W_{h,j}\ge 0,
  \\
  \forall(h',j')\in\cE_{-},\quad
  &W_{h',j'}\le 0, 
\end{tsaligned}
with two exclusive set
$\cE_{+}$ and $\cE_{-}$
such that
\begin{tsaligned}
\cE_{+}\cup\cE_{-}\subseteq\{(h,j)\in\bN^{2}\,|\,h\in[1,d], j\in[1,m]\}. 
\end{tsaligned}
Introducing $\vC\in\{0,\pm1\}^{d\times m}$ as
\begin{tsaligned}
  C_{h,j}
  :=
  \begin{cases}
    +1 \quad&\text{for }(h,j)\in\cE_{+},
    \\
    -1 \quad&\text{for }(h,j)\in\cE_{-},
    \\
    0 \quad&\text{o.w. }
  \end{cases}
\end{tsaligned}
the feasible region can be expressed as
\begin{tsaligned}
  \cS^{\text{m}} :=
  \left\{
  \vW\in\bR^{d\times m}\,\middle|\,
  \vC\odot\vW \ge \vO_{d\times m}
  \right\}. 
\end{tsaligned}
The goal is here to develop algorithms
that find
\begin{tsaligned}
  \vW_{\star}:=\arg\min_{\vW\in\cS^{\text{m}}}P^{\text{m}}(\vW). 
\end{tsaligned}
Define
$\Phi^{\text{m}}(\cdot\,;\,\cA):\bR^{m\times k}\to\bR$
as
\begin{tsaligned}
  \Phi^{\text{m}}(\vS_{\cA}\,;\,\cA)
  :=
  \sum_{i\in\cA}\phi^{\text{m}}_{i}(\vs_{i})
\end{tsaligned}
where $\vS_{\cA}$ is the horizontal concatenation of
columns in $\vS:=\left[\vs_{1},\dots,\vs_{n}\right]\in\bR^{m\times n}$
selected by a minibatch $\cA$.
We here use the following assumptions: 
$\Phi^{\text{m}}(\cdot)$ is a convex function;  
$\Phi^{\text{m}}(\vO)\le n\rloss$;  
$\forall \vS\in\bR^{m\times n}$, $\Phi^{\text{m}}(\vS)\ge 0$; 
$\forall i$, $\lVert\x_{i}\rVert \le R$. 

By extending Algorithm~\ref{alg:srpega-mbtch}, an algorithm for minimization of $P^{\text{m}}(\vW)$ subject to the sign constraints can be developed as described in Algorithm~\ref{alg:srpega-mbtch-mc}. 
\begin{algorithm}[t!]
\caption{
Sign-Constrained Pegasos for Multiclass Classification. 
\label{alg:srpega-mbtch-mc}}
\begin{algorithmic}[1]
  \REQUIRE
  Data matrix $\X\in\bR^{d\times n}$,
  loss function $\Phi^{\text{m}}:\bR^{m\times n}\to\bR$,
  regularization parameter $\lambda\in\bR$,
  sign constraint parameter $\vC\in\{0,\pm 1\}^{d\times m}$, and
  mini-batch size $k$. 
  \STATE \textbf{begin}
  \STATE $\vW_{1}:=\0_{d}$;  \{Initialization\}
  \FOR{$t:=1,\dots,T$}
  \STATE Choose $\cA_{t}\subseteq\{1,\dots,n\}$ uniformly
  at random such that $|A_{t}|=k$. 
  \STATE
  $\Z_{t}:=\vW_{t-1}^\top\X_{\cA_{t}}$; 
  \STATE
  $\vW_{t+1/3}
  :=
  \frac{t-1}{t}\vW_{t} -
  \frac{1}{\lambda t}\X_{\cA_{t}}\left(\nabla\Phi(\Z_{t}\,;\,\cA_{t})\right)^\top$; 
  \STATE
  $\vW_{t+2/3}
  :=
  \vW_{t+1/3}
  + \vC\odot\max(\vO,-\vC\odot\vW_{t+1/3})$; 
  \STATE
  $\vW_{t+1}
  :=
  \min\left(
  1, \frac{\rloss}{\sqrt{\lambda}\lVert\vW_{t+2/3}\rVert_{\text{F}}}
  \right)
  \vW_{t+2/3}$; 
  \ENDFOR
  \STATE \textbf{return }  $\tilde{\vW}:=\sum_{t=1}^{T}\vW_{t}/T$; 
  \STATE \textbf{end.}
\end{algorithmic}
\end{algorithm}
%

The SDCA-based learning algorithm can also be developed for the multiclass classification task. In the algorithm, the dual variables are represented as a matrix $\vA:=\left[\valph_{1},\dots,\valph_{n}\right]\in\bR^{m\times n}$. At each iterate $t$, one of $n$ columns, $\valph_{i}$, is chosen at random instead of choosing one of a dual variable to update the matrix as $\vA^{(t)}:= \vA^{(t-1)}+\Delta\valph\ve_{i}^\top$ where we have used the iterate number $(t)$ as the superscript of $\vA$. To determine the value of $\Delta\valph$, the following auxiliary funcition is introduced:
\begin{multline}
  D_{t}(\Delta \valph\,;\,\vW)
  := 
  -\frac{\lVert\x_{i}\rVert^{2}}{2\lambda^{2}n}
  \lVert\Delta\valph\rVert^{2}
  \\
  -\left<\vW^\top
  \x_{i}, 
  \Delta\valph\right>
  -\phi^{*}_{i}(-\valph^{(t-1)}_{i}-\Delta \valph). 
\end{multline}

\begin{algorithm}[t!]
\caption{
Sign-Constrained SDCA for Multiclass Classification. 
\label{alg:srsdca-mc}}
\begin{algorithmic}[1]
  \REQUIRE
  Data matrix $\X\in\bR^{d\times n}$,
  loss function $\Phi:\bR^{m\times n}\to\bR$,
  regularization parameter $\lambda\in\bR$, and
  sign constraint parameter $\vC\in\{\pm 1,0\}^{d\times m}$. 
  \STATE \textbf{begin}
  \STATE $\vA^{(0)}:=\vO$; $\bar{\vW}^{(0)}:=\vO$; $\vW^{(0)}:=\vO$;  \qquad \{Initialization\}
  \FOR{$t:=1,\dots,T$}
  \STATE
  $\Delta \valph \in \argmax_{\Delta\valph\in\bR^{m}}D_{t}(\Delta\valph\,;\,\vW^{(t-1)})$; 
  \STATE
  $\bar{\vW}^{(t)}:=\bar{\vW}^{(t-1)} + \frac{1}{\lambda n}\x_{i}\Delta\valph^\top$; 
  \STATE
  $\vW^{(t)}
  :=
  \bar{\vW}^{(t)}
  + \vC\odot\max(\vO,-\vC\odot\bar{\vW}^{(t)})$; 
  \ENDFOR
  \STATE
  \qquad \textbf{return } $\tilde{\vW}:=\frac{1}{T-T_{0}}\sum_{t=T_{0}+1}^{T}\vW^{(t-1)}$;  
  \STATE \textbf{end.}
\end{algorithmic}
\end{algorithm}


For both algorithms (Algorithms~\ref{alg:srpega-mbtch-mc} and \ref{alg:srsdca-mc}), we can bound the required number of iterations similar to those presented in Theorems~\ref{thm:srpega-bound}
and \ref{thm:converg-srsdca}.

\section{Experiments}
In this section, experimental results are reported in order to illustrate the effects of the sign constraints on classification and to demonstrate the convergence behavior. 

\subsection{Prediction Performance}
The pattern recognition performance of the sign-constrained learning was
examined on two tasks: \textit{Escherichia coli} (\textit{E.~coli}) prediction and protein function prediction.

\paragraph*{\textit{E.~coli} Prediction}
The first task is to predict \textit{E.~coli} counts in river water. The \textit{E.~coli} count has been used as an indicator for fecal contamination in water environment in many parts of the world~\citep{ScoRosJen02-aem}. In this experiment, the data points with \textit{E.~coli} counts over 500 most probable number (MPN)/100 mL are assigned to positive class, and the others are negative. The hydrological and water quality monitoring data are used for predicting \textit{E.~coli} counts to be positive or negative.

For ensuring the microbial safety in water usage, it is meaningful to predict \textit{E.~coli} counts on a real-time basis. The concentration of \textit{E.~coli} in water, which is measured by culture-dependent methods~\citep{KobSanHat13-amb}, has been used to monitor the fecal contamination in water environment, and has been proved to be effective to prevent waterborne infectious diseases in varied water usage styles. On the other hand, the real-time monitoring of \textit{E.~coli} counts has not yet been achieved. It take at least ten hours to obtain \textit{E.~coli} counts by culture-dependent methods, and also at least several hours are needed to measure the concentration of \textit{E.~coli} by culture-independent methods~\citep{IshNakOza14-est,IshKitSeg14-aem}, such as polymerase chain reaction. Since it is possible to measure the some of the hydrological and water quality data with real-time sensors, the real-time prediction of \textit{E.~coli} counts will be realized if the hydrological and water quality data are available for the \textit{E.~coli} count prediction. 

Many training examples are required to obtain a better generalization performance.
A serious issue, however, is that measuring the concentration of \textit{E. Coli} is time-consuming and the cost of reagents is expensive.
We here demonstrate that this issue can be relaxed by exploiting the domain knowledge hoarded in the field of water engineering.

The hydrological and water quality data contain nine explanatory variables, WT, pH, EC, SS, DO, BOD, TN, TP, and flow rate. 
The explanatory variable $pH$ is divided into two variables, $\text{pH}_{+}\leftarrow\max(0,\text{pH}-7)$ and $\text{pH}_{-}\leftarrow\max(0,7-\text{pH})$.
It is well-known, in the field of water engineering, that 
\textit{E.~coli} is increased, as WT, EC, SS, BOD, TN, and TP are larger, 
and
as $\text{pH}_{+}$, $\text{pH}_{-}$，DO, and the flow rate are smaller.
From this fact, we restrict the sign of entries in the predictor parameter $\vw$ as follows.
\begin{itemize}
\item Coefficients $w_{h}$ of six explanatory variables, WT, EC, SS, BOD, TN, and TP must be non-negative.
\item Coefficients $w_{h}$ of four explanatory variables, $\text{pH}_{+}$, $\text{pH}_{-}$, DO, flow rate must be non-positive.
\end{itemize}
We actually measured the concentrations of \textit{E.~coli} 177 times from December 5th, 2011 to April 17th, 2013. We obtained 177 data points including 88 positives and 89 negatives. We chose ten examples out of 177 data points at random to use them for training, and the other 167 examples were used for testing. The prediction performance is evaluated by the precision recall break-even point (PRBEP)~\citep{Joachims05-icml} and the ROC score. We compared the classical SVM with the sign-constrained SVM (SC-SVM) to examine the effects of sign constraints. We repeated this procedure 10,000 times and obtained 10,000 PRBEP and 10,000 ROC scores.

\begin{figure}[t!]
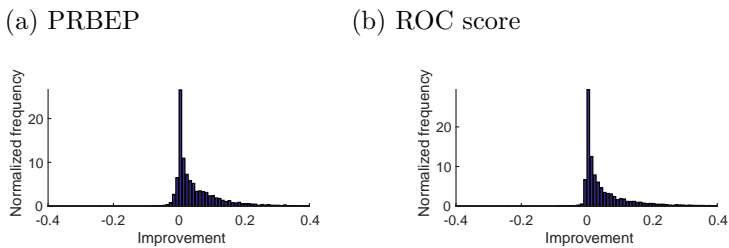

  \begin{center}
    \begin{tabular}{ll}
      (a) PRBEP & 
      (b) ROC score \\ \\
      \includegraphics[width=0.22\textwidth]{001-k.demo332_02.prbep.eps}
      &
      $\qquad$
      \includegraphics[width=0.22\textwidth]{002-k.demo332_02.roc.eps}
    \end{tabular}
  \end{center}
  \caption{
    Improvements of generalization performances on \textit{E. Coli} prediction. 
    \label{fig:prot404-demo332-mizu}}
\end{figure}

SC-SVM achieved significant improvement compared to the classical SVM. SC-SVM achieved PRBEP and ROC score of 0.808 and 0.863 on average over 10,000 trials, whereas those of the classical SVM were 0.757 and 0.810, respectively. The difference from the classical SVM on each trial is plotted in the histograms of Figure~\ref{fig:prot404-demo332-mizu}. Positive improvements of ROC scores were obtained in 8,932 trials out of 10,000 trials, whereas ROC scores were decreased only for 796 trials. 
For PRBEP, positive improvements were obtained on 7,349 trials, whereas deteriorations were observed only on 1,069 trials.
\begin{figure*}[t!]
  \begin{center}
    \begin{tabular}{lll}
      (a) Covtype &
      (b) W8a &
      (c) Phishing
      \\
      \includegraphics[width=0.3\textwidth]{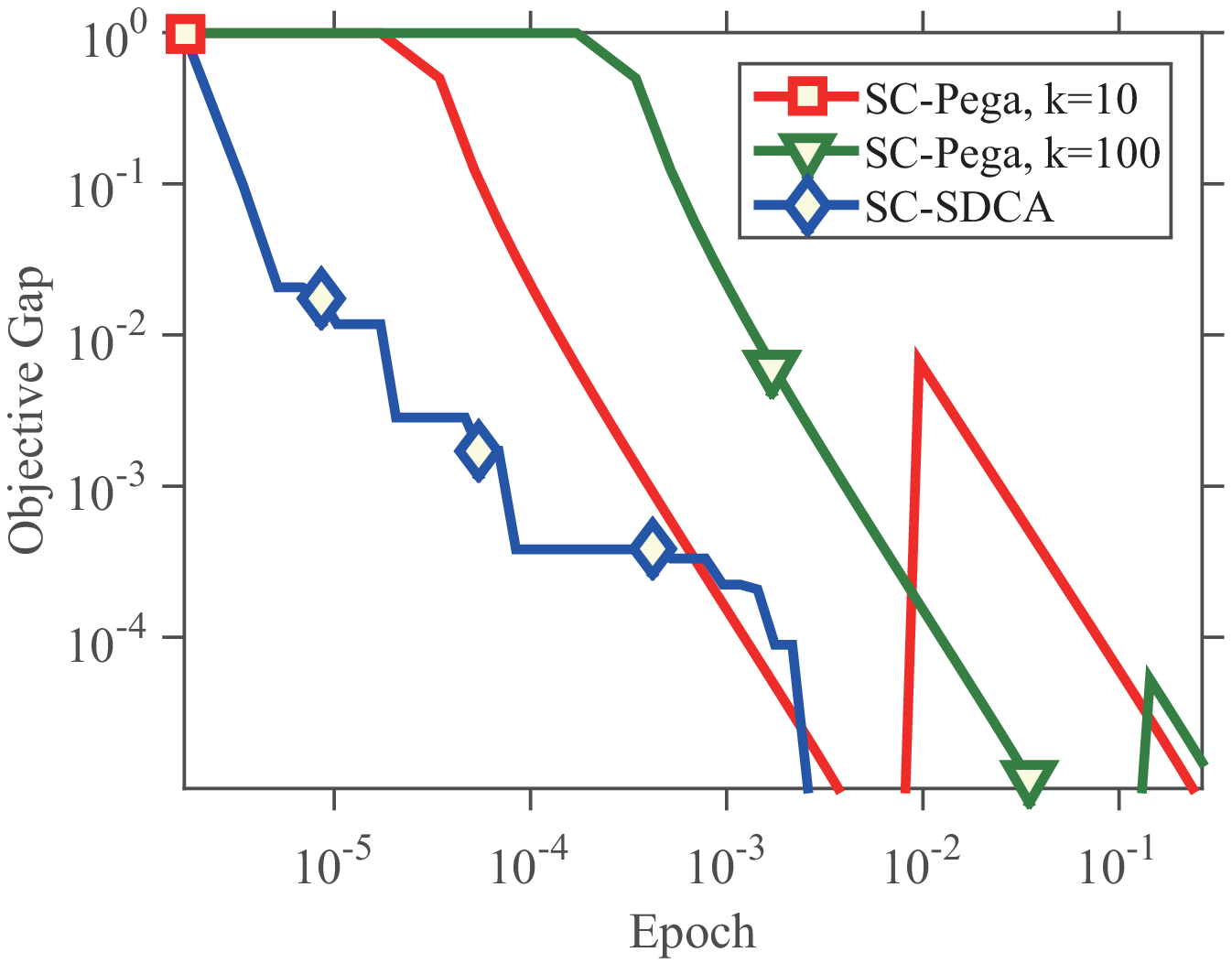} &
      \includegraphics[width=0.3\textwidth]{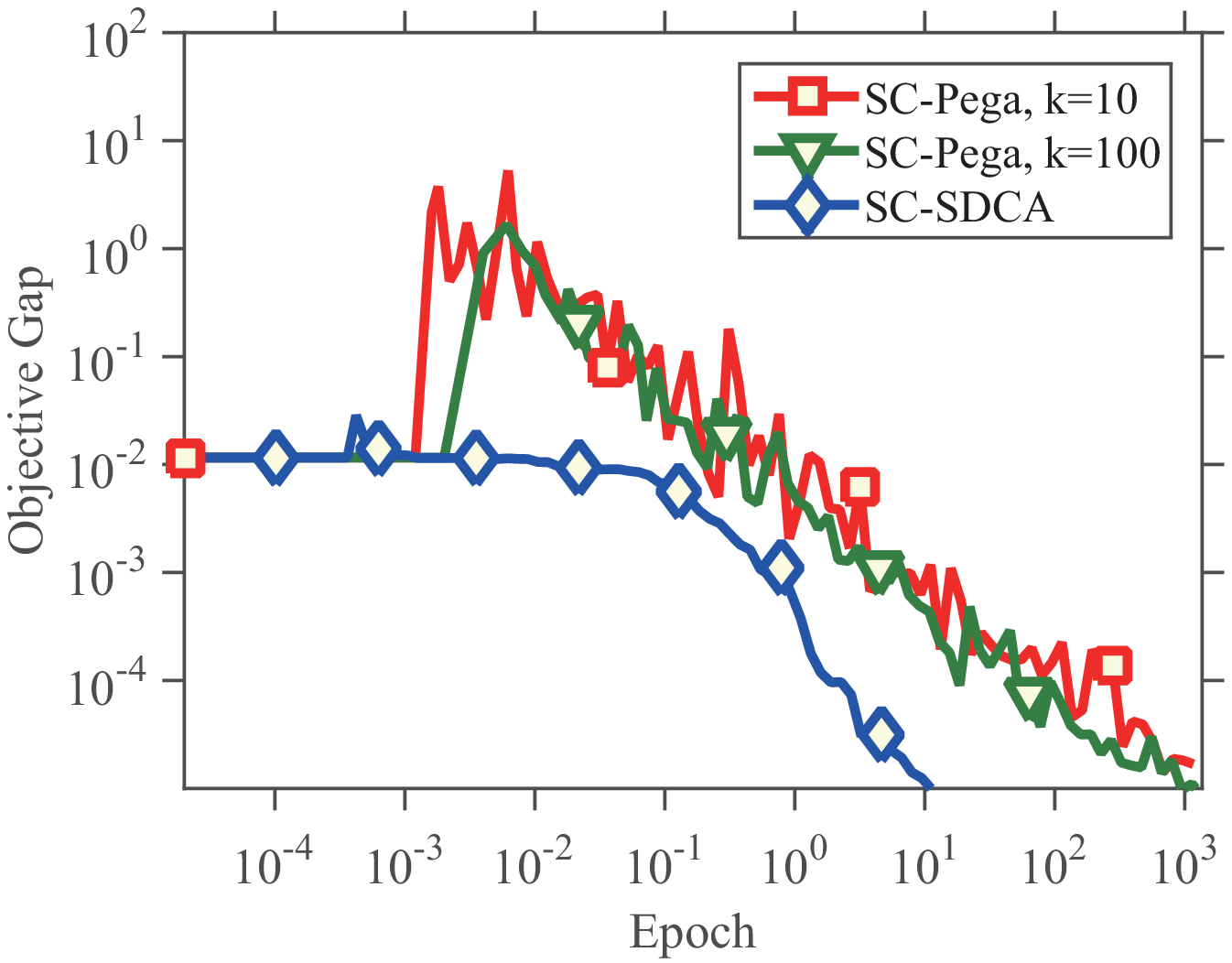}
      &
      \includegraphics[width=0.3\textwidth]{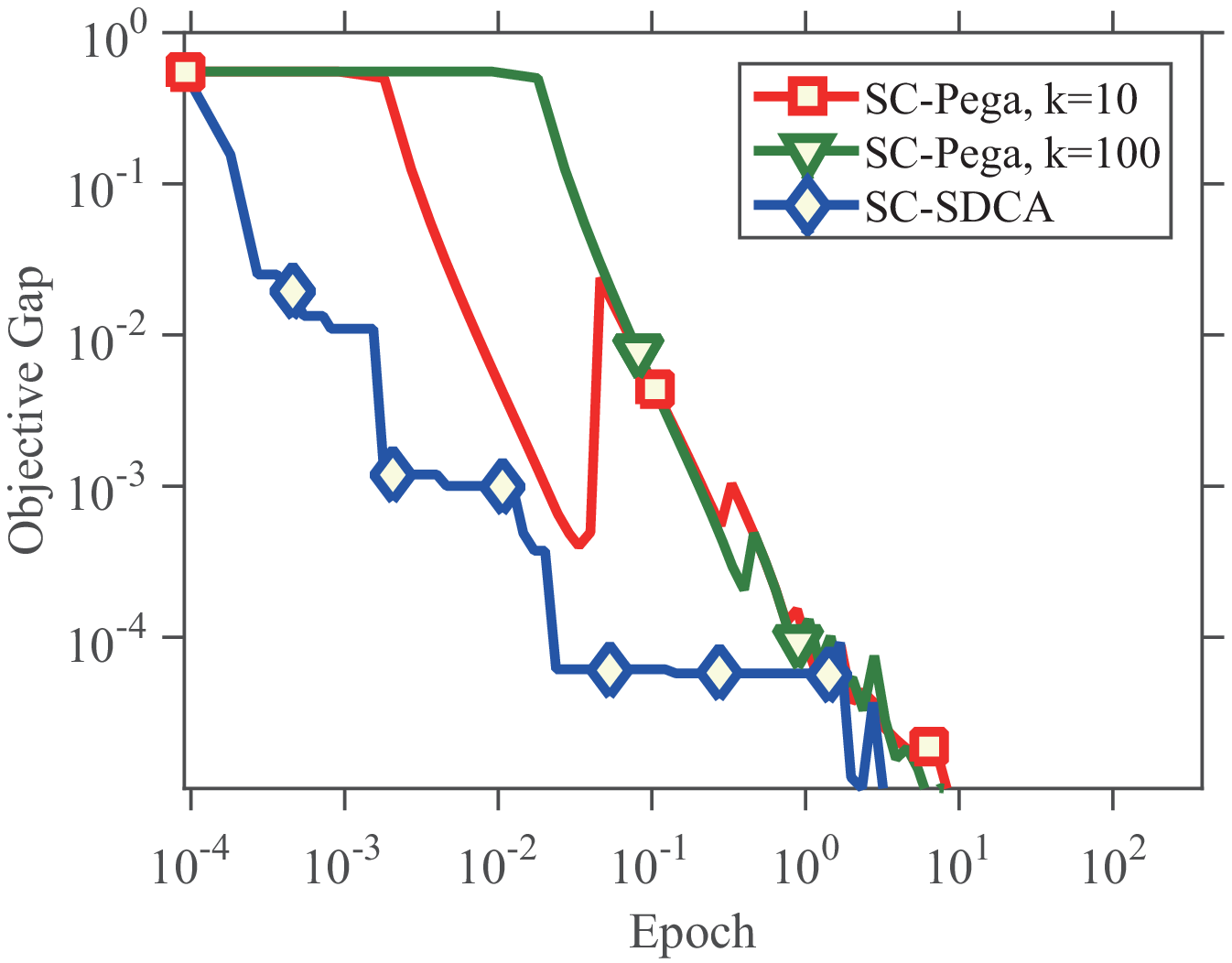}
      \\
    \end{tabular}
  \end{center}
  \caption{
    Comparison of different optimization methods. 
    \label{fig:demo423-gap}}
\end{figure*}
\paragraph*{Protein Function Prediction}
\begin{table}[t!]
  \begin{center}
    \caption{ROC Scores for protein function prediction. \label{tab:demo383}}
    \begin{tabular}{|c|cc|}
      \hline
      Category & SC-SVM & SVM\\
      \hline
1 & \textbf{0.751} (0.011)  & 0.730 (0.010) \\
2 & \textbf{0.740} (0.016)  & 0.680 (0.015) \\
3 & \textbf{0.753} (0.011)  & 0.721 (0.011) \\
4 & \textbf{0.762} (0.010)  & 0.734 (0.010) \\
5 & \textbf{0.769} (0.012)  & 0.691 (0.013) \\
6 & \textbf{0.690} (0.014)  & 0.614 (0.014) \\
7 & \textbf{0.713} (0.024)  & 0.618 (0.022) \\
8 & \textbf{0.725} (0.019)  & 0.667 (0.019) \\
9 & \textbf{0.655} (0.024)  & 0.578 (0.023) \\
10 & \textbf{0.743} (0.016)  & 0.710 (0.014) \\
11 & \textbf{0.535} (0.019)  & 0.492 (0.018) \\
12 & \textbf{0.912} (0.011)  & 0.901 (0.011) \\
\hline
\end{tabular}
  \end{center}
\end{table}
In the field of molecular biology, understanding the functions of proteins is positioned as a key step for elucidation of cellular mechanisms. Sequence similarities have been a major mean to predict the function of an unannotated protein. At the beginning of this century, the prediction accuracy has been improved by combining sequence similarities with discriminative learning. The method, named SVM-Pairwise~\citep{LiaNob03-jcb}, uses a feature vector that contains pairwise similarities to annotated protein sequences. Several other literature~\citep{LiuZhaXu14,OguMum06,LanBieCri04,LanDenCri04} have also provided empirical evidences for the fact that the SVM-Pairwise approach is a powerful framework.  Basically, if $n$ proteins are in a training dataset, the feature vector has $n$ entries, $x_{1}, \dots, x_{n}$. If we suppose that the first $n_{+}$ proteins in the training set are in positive class and the rest are negative, then the first $n_{+}$ similarities $x_{1}, \dots, x_{n_{+}}$ are sequence similarities to positive examples, and $x_{n_{+}+1}, \dots, x_{n}$ are similarities to negative examples. The $n$-dimensional vectors are fed to SVM and get the weight coefficients $\vw:=\left[w_{1},\dots,w_{n}\right]^\top$. Then, the prediction score of the target protein is expressed as
\begin{tsaligned}
  \sum_{i=1}^{n_{+}}w_{i}x_{i}
  +
  \sum_{i'=n_{+}+1}^{n}w_{i}x_{i}. 
\end{tsaligned}
The input protein sequence is predicted to have some particular cellular function if the score is over a threshold. 
It should be preferable that the first $n_{+}$ weight coefficients $w_{1},\dots,w_{n_{+}}$ are non-negative and that the rest of $(n-n_{+})$ weight coefficients $w_{n_{+}+1},\dots,w_{n}$ are non-positive. The SVM-Pairwise approach does not ensure those requirements. Meanwhile, our approach is capable to explicitly impose the constraints of
\begin{tsaligned}
  w_{1}\ge 0,\dots,w_{n_{+}}\ge 0,
\quad\text{and}\quad
  w_{n_{+}+1}\le 0,\dots,w_{n}\le 0. 
\end{tsaligned}

This approach was applied to predict protein functions in \textit{Saccharomyces cerevisiae} (\textit{S. cerevisiae}). The annotations of the protein functions are provided in MIPS Comprehensive Yeast Genome Database (CYGD). The dataset contains 3,583 proteins. The Smith-Waterman similarities available from \url{https://noble.gs.washington.edu/proj/sdp-svm/} were used as sequence similarities among the proteins. The number of categories was 12. Some proteins have multiple cellular functions. Indeed, 1,343 proteins in the dataset have more than one function. From this reason, we pose 12 independent binary classification tasks instead of a single multi-class classification task. 3,583 proteins were randomly splited in half to get two datasets. The one was used for training, and the other was for testing. For 12 classification tasks, we repeated this procedure 100 times, and we obtained 100 ROC scores. 

Table~\ref{tab:demo383} reports the ROC scores averaged over 100 trials and the standard deviations for 12 binary classification tasks. The sign constraints significantly surpassed the classical training for all 12 tasks. Surprisingly, we observed that the ROC score of SC-SVM is larger than that of the classical SVM in every trial. 

\subsection{Convergence}
We carried out empirical evaluation of the proposed optimization methods, the sign-constrained Pegasos (SC-Pega) and the sign-constrained SDCA (SC-SDCA), in order to illustrate the convergence of our algorithms to the optimum. For SC-Pega, we set the mini-batch size to $k=10$ and $k=100$. In this experiments, we used the smoothed hinge loss with $\gamma=0.01$ and $\lambda=1/n$. We used three datasets, Covtype ($n=581,012$ and $d=54$), W8a ($n=49,749$ and $d=300$),
and Phishing ($n=11,055$ and $d=68$). The three datasets are for binary classification and available from LIBSVM web site (https://www.csie.ntu.edu.tw/~cjlin/libsvmtools/datasets/). 

Figure~\ref{fig:demo423-gap} depicts the primal objective gap against epochs, where the primal objective gap is defined as $P(\vw)-P(\vw_{\star})$.
As expected in theoretical results, SC-SDCA converged to the optimum faster than SC-Pega except on the dataset Phishing. No significant difference between different mini-batch sizes is observed. 
\section{Conclusions}
In this paper, we presented two new algorithms for minimizing regularized empirical loss subject to sign constraints. The two algorithms are based on Pegasos and SDCA, both of which have a solid theoretical support for convergence. The sign-constrained versions, named SC-Pega and SC-SDCA, respectively, enjoy the same convergence rate as the corresponding original algorithms. The algorithms were demonstrated in two applications. The one is posing sign constraints according to domain knowledge, and the other is improving the SVM-Pairwise method by sign constraints.

\section*{Acknowledgements} %
TK was supported by JPSP KAKENHI Grant Number 26249075 and 40401236.

\bibliographystyle{natbib}

\setcounter{section}{0}
\renewcommand{\thesection}{\Alph{section}}   
\section{Proofs and Derivations}

\subsection{Proof of Theorem~\ref{thm:srpega-bound}}
  \label{ss:proof-thm:srpega-bound}
\citet{Shalev-Shwartz11-pegasos} have used the following lemma, given below, to obtain the bound.
\begin{lemma-waku}[\citet{Hazan07-Logarithmic}]\label{lem:pega-lem01}
  Let $f_{1},\dots,f_{T}$ be a sequence of
  $\lambda$-strongly convex functions.
  Let $\cC$ be a closed convex set and
  define $\Pi_{\cC}(\vw):=\argmin_{\vw'\in\cC}\|\vw'-\vw\|$.
  Let $\vw_{1},\dots,\vw_{T+1}$ be a sequence of vectors
  such that $\vw_{1}\in\cC$ and for $t\ge 1$,
  $\vw_{t+1}:=\Pi_{\cC}(\vw_{t}-\nabla_{t}/(\lambda t))$,
  where $\nabla_{t}\in\partial f_{t}(\vw_{t})$. 
  Assume that $\forall t\in\bN$, $\|\nabla_{t}\|\le G$.
  Then, for $\forall \vu\in\cC$, it holds that
  \begin{align}
    \frac{1}{T}\sum_{t=1}^{T}f_{t}(\vw_{t})
    \le
    \frac{1}{T}\sum_{t=1}^{T}f_{t}(\vu)
    +
    \frac{(1+\log(T))G^{2}}{2\lambda T}. 
  \end{align}
\end{lemma-waku}
We, too, have used Lemma~\ref{lem:pega-lem01}
to obtain Theorem~\ref{thm:srpega-bound}
for our sign-constrained learning problem~\eqref{eq:prob-rlm-signcon}.
To this end, we find the following lemma. 
\begin{lemma-waku}\label{lem:hazen-applicable}
  Let $\cB$ be $\rloss/\sqrt{\lambda}$-ball defined as
  \begin{tsaligned}
    \cB:=
    \left\{\vw\in\bR^{d}\,\middle|\,
    \lVert\vw\rVert\le \sqrt{\frac{\rloss}{\lambda}}\right\}.
  \end{tsaligned}
  and $\cS$ be the set defined in \eqref{eq:fearegion}.
  Then, the intersection of the two sets are
  closed and convex. It holds that
  \begin{tsaligned}\label{eq:update-lem:hazen-applicable}
    \vw_{t+1} = \Pi_{\cB\cap\cS}
    \left(\vw_{t}-\frac{1}{\lambda t}\nabla P_{t}(\vw_{t})
    \right)
  \end{tsaligned}
  for $\forall t\in\bN$.  
  Furthermore, 
  the optimal solution $\vw_{\star} := \argmin_{\vw\in\cS}P(\vw)$
  is in the intersection of the two sets. Namely, 
  \begin{tsaligned}\label{eq:opt-lem:hazen-applicable}
    \vw_{\star}\in\cB\cap\cS. 
  \end{tsaligned}
\end{lemma-waku}
See Subsection~\ref{ss:proof-lem:hazen-applicable} for proof
of Lemma~\ref{lem:hazen-applicable}. 
The above lemma suggests that  
the setting of $f_{t}:=P_{t}$, $\cC:=\cB\cap\cS$
and $\vu:=\vw_{\star}$
fulfills the assumptions of Lemma~\ref{lem:pega-lem01}.
An upper bound of the norm of the gradient of $f_{t}$
is given by
\begin{tsaligned}\label{eq:ub-for-setting-G-in-pega}
  \left\lVert\nabla f_{t}(\vw_{t})\right\rVert
  = \left\lVert\nabla P_{t}(\vw_{t})\right\rVert
  \le
  \sqrt{\rloss \lambda} + LR. 
\end{tsaligned}
See Subsection~\ref{ss:deriv-eq:ub-for-setting-G-in-pega}
for derivation of \eqref{eq:ub-for-setting-G-in-pega}. 
By setting $G=\sqrt{\rloss \lambda} + LR$, 
Theorem~\ref{thm:srpega-bound} is established.
\qed

\subsection{Proof of Lemma~\ref{lem:hazen-applicable}}
\label{ss:proof-lem:hazen-applicable}
Lemma~\ref{lem:hazen-applicable}
states the following three claims.
\begin{itemize}
\item $\cB\cap\cS$ is a closed and convex set.
\item \eqref{eq:update-lem:hazen-applicable} is hold.
\item \eqref{eq:opt-lem:hazen-applicable} is hold.
\end{itemize}
Apparently, $\cB\cap\cS$ is a closed and convex set
because the both sets are closed and convex.
We shall show \eqref{eq:update-lem:hazen-applicable}
and then \eqref{eq:opt-lem:hazen-applicable}.

\paragraph*{Proof of \eqref{eq:update-lem:hazen-applicable}}
To prove \eqref{eq:update-lem:hazen-applicable},
it suffices to show the projection from a point $\z\in\bR^{d}$
onto the set $\cB\cap\cS$ is given by
\begin{tsaligned}\label{eq:update-prime-lem:hazen-applicable}
  \Pi_{\cB\cap\cS}(\z)
  =
  \min\left\{
  1, \sqrt{\frac{\rloss}{\lambda\left\lVert\Pi_{\cS}(\z)\right\rVert^{2}}}
  \right\}
  \Pi_{\cS}(\z). 
\end{tsaligned}
The projection problem can be expressed as
\begin{tsaligned}
  \text{min }\quad&
  \frac{1}{2}\|\x-\z\|^{2}
  \qquad
  \text{wrt }\quad \x\in\bR^{d}
  \\
  \text{subject to }\quad&
  \|\x\|^{2}\le \frac{\rloss}{\lambda}, \quad
  \vc\odot\x \ge \0_{d}. 
\end{tsaligned}
With non-negative
dual variables $\vbeta\in\bR^{d}_{+}$ and $\eta\in\bR_{+}$,
the Lagrangian function is given by
\begin{multline}
  \cL_{\cB\cap\cS}(\x,\vbeta,\eta)
  :=
  \frac{1}{2}\|\x-\z\|^{2}
  \\
  -
  \left<\vbeta,\vc\odot\x\right>
  +
  \frac{\eta}{2}
  \left(
  \|\x\|^{2}
  -
  \frac{\rloss}{\lambda}
  \right). 
\end{multline}
Let $(\x_{\star},\vbeta_{\star},\eta_{\star})$ be the saddle
point of $\min_{\x}\max_{\vbeta,\eta} \cL_{\cB\cap\cS}(\x,\vbeta,\eta)$.
Then, $\x_{\star} = \Pi_{\cB\cap\cS}(\z)$.
At the saddle point, it holds that
$\nabla_{\x}  \cL_{\cB\cap\cS}=\0$, yielding
\begin{tsaligned}\label{eq:01-proof-of-lem:pegaproj-onto-B-cap-S}
  \x = \frac{1}{\eta+1}(\z+\vc\odot\vbeta). 
\end{tsaligned}
The dual objective is written as
\begin{tsaligned}
  \cD_{\cB\cap\cS}(\vbeta,\eta)
  &= \min_{\x}\cL_{\cB\cap\cS}(\x,\vbeta,\eta)
  \\
  &= \cL_{\cB\cap\cS}(\frac{1}{\eta+1}(\z+\vc\odot\vbeta),\vbeta,\eta)
  \\
  &= -\frac{1}{2(\eta+1)}\|\z+\vc\odot\vbeta\|^{2}
  + \frac{\eta}{2\lambda} + \frac{1}{2}\|\z\|^{2}. 
\end{tsaligned}
This implies that $\cD_{\cB\cap\cS}(\cdot,\cdot)$ is maximized when
$\vbeta = (-\vc\odot\z)_{+}$. Note that this does not
depend on the value of $\eta$.
Substituting this into
\eqref{eq:01-proof-of-lem:pegaproj-onto-B-cap-S},
we have
\begin{tsaligned}\label{eq:02-proof-of-lem:pegaproj-onto-B-cap-S}
  \x = \frac{1}{\eta+1}(\z+\vc\odot(-\vc\odot\z)_{+})
  = \frac{1}{\eta+1}\Pi_{\cS}(\z), 
\end{tsaligned}
where the last equality follows from the fast
that $\Pi_{\cS}(\z)=\z+\vc\odot(-\vc\odot\z)_{+}$
which can be shown as follows.
The Lagrangian function for the problem of
projection of $\z$ onto $\cS$ is given by
$\cL_{\cS}(\x,\vbeta) = \cL_{\cB\cap\cS}(\x,\vbeta,0)$,
and, with a similar derivation, the dual objective
is $\cD_{\cS}(\vbeta) = \cD_{\cB\cap\cS}(\vbeta,0)$
which is maximized at $\vbeta = (-\vc\odot\z)_{+}$
yielding $\Pi_{\cS}(\z) = \z+(-\vc\odot\z)_{+}$. 

Next, we find the optimal $\eta$.
The dual objective is
\begin{tsaligned}
  \cD_{\cB\cap\cS}((-\vc\odot\z)_{+},\eta)
  =
  -\frac{1}{2}\|\Pi_{\cS}(\z)\|^{2}(\eta+1)^{-1}
  - \frac{\rloss}{2\lambda}\eta + \frac{1}{2}\|\z\|^{2}
\end{tsaligned}
with the derivative
\begin{tsaligned}
  \nabla_{\eta}\cD_{\cB\cap\cS}((-\vc\odot\z)_{+},\eta)
  =
  \frac{1}{2}\|\Pi_{\cS}(\z)\|^{2}(\eta+1)^{-2}
  - \frac{\rloss}{2\lambda}. 
\end{tsaligned}
Setting the derivative to zero and noting that
$\eta$ is a non-negative variable, we get
\begin{tsaligned}
  \eta_{\star}
  =
  \max\left( 0, \sqrt{\frac{\lambda}{\rloss}}
  \left\lVert\Pi_{\cS}(\z)\right\rVert-1 \right). 
\end{tsaligned}
Substituting this into
\eqref{eq:02-proof-of-lem:pegaproj-onto-B-cap-S},
we obtain
\begin{multline}
  \Pi_{\cB\cap\cS}(\z)
  =
  \x_{\star}
  =
  \frac{1}{1+\max( 0, \sqrt{\lambda/\rloss}
    \left\lVert\Pi_{\cS}(\z)\right\rVert-1 )}
  \Pi_{\cS}(\z)
  \\
  =
  \min\left\{1,\frac{\sqrt{\rloss}}{\sqrt{\lambda}
     \left\lVert\Pi_{\cS}(\z)\right\rVert}\right\}\Pi_{\cS}(\z). 
\end{multline}
Thus, \eqref{eq:update-prime-lem:hazen-applicable} is established. 

\paragraph*{Proof of \eqref{eq:opt-lem:hazen-applicable}}
We use the following problem dual to \eqref{eq:prob-rlm-signcon}: 
\begin{tsaligned}\label{eq:dual-rlm-signcon}
  \text{max}\quad
  -\frac{\lambda}{2}
  \left\lVert\Pi_{\cS}\left(\frac{\X\valph}{\lambda n}\right)\right\rVert^{2}
  -\frac{1}{n}\Phi^{*}(-\valph)
  \quad\text{wrt}\quad \valph\in\bR^{n}. 
\end{tsaligned}
Let $\valph_{\star}$ be the solution optimal to
the dual problem~\eqref{eq:dual-rlm-signcon}.
The primal optimal solution can be recovered by
\begin{tsaligned}\label{eq:wst-recovery-from-alphst}
  \vw_{\star} = \Pi_{\cS}\left(\frac{1}{\lambda n}\X\valph_{\star}\right)
\end{tsaligned}
with no duality gap. The loss term in the objective
of the dual problem is bounded from above as
\begin{tsaligned}\label{eq:loss-in-dual-bounded}
  -\frac{1}{n}\Phi^{*}(-\valph)
  &=
  -\frac{1}{n}\max_{\vs\in\bR^{n}}
  \left(
  \left<\vs,-\valph\right> - \Phi(\vs)
  \right)
  \\
  &
  =
  \frac{1}{n}\min_{\vs\in\bR^{n}}
  \left(
  \left<\vs,\valph\right> + \Phi(\vs)
  \right)
  \\
  &=
  \frac{1}{n}
  \left(
  \left<\0,\valph\right> + \Phi(\0)
  \right)
  =
  \frac{1}{n}
  \Phi(\0) \le \rloss. 
\end{tsaligned}
The square norm of the primal optimal solution is bounded as
\begin{tsaligned}
  &\lVert\vw_{\star}\rVert^{2}
  =
  \frac{1}{2}\lVert\vw_{\star}\rVert^{2}
  +
  \frac{1}{2}\lVert\vw_{\star}\rVert^{2}
  \\
  &\le
  \frac{1}{2}\lVert\vw_{\star}\rVert^{2}
  +
  \frac{1}{\lambda}
  \left(
  \frac{\lambda}{2}\lVert\vw_{\star}\rVert^{2}
  +
  \frac{1}{n}\Phi(\X^\top\vw_{\star})
  \right)
  \\
  &=
  \frac{1}{2}\lVert\vw_{\star}\rVert^{2}
  +
  \frac{1}{\lambda}P(\vw_{\star})
  \\
  &=
  \frac{1}{2}\lVert\vw_{\star}\rVert^{2}
  +
  \frac{1}{\lambda}
  \left(
  -\frac{\lambda}{2}
  \left\lVert\Pi_{\cS}\left(\frac{\X\valph_{\star}}{\lambda n}\right)\right\rVert^{2}
  -\frac{1}{n}\Phi^{*}(-\valph_{\star})
  \right)
  \\
  &=
  \frac{1}{2}\lVert\vw_{\star}\rVert^{2}
  -
  \frac{1}{\lambda}
  \left(
  \frac{\lambda}{2}
  \left\lVert\vw_{\star}\right\rVert^{2}
  +
  \frac{1}{n}\Phi^{*}(-\valph_{\star})
  \right)
  \\
  &=
  -\frac{1}{\lambda n}\Phi^{*}(-\valph_{\star})
  \le
  \frac{\rloss}{\lambda} 
\end{tsaligned}
where the first inequality, 
the third and fourth equalities,
and the last inequality
follow
from Assumption~\ref{assum:four-for-rlm-signcon}(c),
no duality gap,
\eqref{eq:wst-recovery-from-alphst}, and 
\eqref{eq:loss-in-dual-bounded}, respectively.
Therefore, $\vw_{\star}\in\cB$. 
Furthermore, $\vw_{\star}$ is feasible so
$\vw_{\star}\in\cS$. 
Hence, \eqref{eq:opt-lem:hazen-applicable} is established. 
\qed

\subsection{Derivation of \eqref{eq:ub-for-setting-G-in-pega}}
\label{ss:deriv-eq:ub-for-setting-G-in-pega}
  This inequality leads to a bound of
  The norm of the gradient of the loss term in $P_{t}(\cdot)$
  can be bounded as 
\begin{multline}
  \left\lVert
  \frac{\partial}{\partial\vw} \Phi(\X_{\cA_{t}}^\top\vw\,;\,\cA_{t})
  \right\rVert
  =
  \left\lVert  
  \sum_{i\in\cA_{t}}\x_{i} \nabla \phi_{i}(\left<\x_{i},\vw\right>)
  \right\rVert
  \\
  \le
  \sum_{i\in\cA_{t}}  \left\lVert \x_{i}  \right\rVert
  \nabla \phi_{i}(\left<\x_{i},\vw\right>)
  \le
  kLR. 
\end{multline}
Using this, \eqref{eq:ub-for-setting-G-in-pega} is
derived as
\begin{tsaligned}
  \left\lVert\nabla f_{t}(\vw_{t})\right\rVert
  &= \left\lVert\nabla P_{t}(\vw_{t})\right\rVert
  \\
  &\le
  \lambda \left\lVert\nabla \vw_{t} \right\rVert
  +
  \frac{1}{k}
  \left\lVert
  \frac{\partial}{\partial\vw} \Phi(\X_{\cA_{t}}^\top\vw\,;\,\cA_{t})
  \right\rVert
  \\
  &\le
  \lambda \sqrt{\frac{\rloss}{\lambda}}
  +
  \frac{kLR}{k} 
  \le
  \sqrt{\rloss \lambda} + LR. 
\end{tsaligned}
\qed


\subsection{Derivation of \eqref{eq:sdca-prox-update}}
\label{ss:deriv-eq:sdca-prox-update}
Exploiting proof techniques used in ProxSDCA
\citep{Shalev-Shwartz2013a-Accelerated}, 
we here limit the form of $\Delta\alpha$ to
$sq$ where $q:=u_{i}-\alpha_{i}^{(t-1)}$ and
$u_{i}\in-\partial\phi_{i}(z_{i})$.

Denote by $D^{0}(\valph)$ the objective function
of the dual problem~\eqref{eq:dual-rlm-signcon}.
Suppose that $i$-th example is chosen at $t$-th iterate.
The new value of the regularization term in $D^{0}(\valph)$
is given by
\begin{multline}
  -\frac{\lambda}{2}
  \left\lVert\Pi_{\cS}\left(\frac{\X(\valph+sq\ve_{i})}{\lambda n}\right)\right\rVert^{2}
  \ge
  -\frac{\lambda}{2}
  \left\lVert\Pi_{\cS}\left(\bar{\vw}^{(t-1)}\right)
  +
  \frac{sq}{\lambda n}\x_{i}\right\rVert^{2}
  \\
  \ge
  -\frac{\lambda}{2}
  \left\lVert\vw^{(t-1)}\right\rVert^{2}
  -
  \frac{s}{n}z_{i}q
  -
  \frac{1}{2\lambda}
  \left(\frac{s}{n}\right)^{2}R^{2}q^{2}. 
\end{multline}
where the first inequality follows from the following inequality
\begin{tsaligned}\label{eq:ub-of-gst-2nd}
    \forall\vv, \forall\vDel\in\bR^{d}, \quad
    \left\lVert\Pi_{\cS}(\vv)+\vDel\right\rVert
    \ge
    \left\lVert\Pi_{\cS}(\vv+\vDel)\right\rVert,  
\end{tsaligned}
and the second inequality is derived from the fact
of $\vw^{(t-1)}=\Pi_{\cS}\left(\bar{\vw}^{(t-1)}\right)$
and the assumption of $\lVert\x_{i}\rVert\le R$. 
We shall prove \eqref{eq:ub-of-gst-2nd}
in Subsection~\ref{ss:deriv-eq:ub-of-gst-2nd}. 

The improvement of the dual objective is expressed as
\begin{tsaligned}
  & D^{0}(\valph^{(t-1)}+sq\ve_{i})-D^{0}(\valph^{(t-1)})
  \\
  &=
  -\frac{\lambda}{2}
  \left\lVert\Pi_{\cS}\left(\frac{\X(\valph+sq\ve_{i})}{\lambda n}\right)\right\rVert^{2}
  +\frac{\lambda}{2}
  \left\lVert\vw^{(t-1)}\right\rVert^{2}
  \\
  &\qquad -\frac{1}{n}\phi_{i}(-\alpha_{i}-sq)
  +\frac{1}{n}\phi_{i}(-\alpha_{i})
  \\
  &\ge
  -
  \frac{1}{2\lambda}
  \left(\frac{s}{n}\right)^{2}R^{2}q^{2}
  +
  \frac{(1-s)s \gamma}{2n}q^{2}
  \\
  &\qquad+
  \frac{s}{n}
  \left(
  z^{(t)}\alpha_{i} + \phi_{i}^{*}(-\alpha_{i})-\phi_{i}^{*}(-u_{i}) - z^{(t)}u 
  \right)
  \\
  &=
  -
  \frac{s^{2}}{2n}q^{2}\gamma s_{\text{lb}}^{-1}
  +
  \frac{s}{n}
  \left(
  z^{(t)}\alpha_{i} + \phi_{i}^{*}(-\alpha_{i}) + \phi_{i}(z^{(t)})
  +\frac{\gamma q^{2}}{2}
  \right)
\end{tsaligned}
Thus, the value of $s$ maximizing the lower-bound can be
given by
\begin{tsaligned}
  s =
  \text{Clip}_{[0,s^{-1}_{\text{lb}}]}
  \left(
  \frac{1}{2} +
  \frac{z^{(t)}\alpha_{i} + \phi_{i}^{*}(-\alpha_{i}) + \phi_{i}(z^{(t)})}{\gamma q^{2}}
  \right)s_{\text{lb}}.
\end{tsaligned}
Thus, \eqref{eq:sdca-prox-update} is derived.
\qed


\subsection{Derivation of \eqref{eq:ub-of-gst-2nd}}
\label{ss:deriv-eq:ub-of-gst-2nd}
\newcommand{\gstar}{{a}}
\newcommand{\barg}{{b}}

For $h=1,\dots,d$, letting
\begin{tsaligned}
  \barg_{h}(\Delta;v)
  :=
  \begin{cases}
    0.5(\Delta+(v)_{+})^{2} &\quad\text{for }h\in\cI_{+},
    \\
    0.5(\Delta+v)^{2} &\quad\text{for }h\in\cI_{0},
    \\
    0.5(\Delta+(-v)_{+})^{2} &\quad\text{for }h\in\cI_{-},    
  \end{cases}
\end{tsaligned}
and
\begin{tsaligned}
  \gstar_{h}(v)
  :=
  \begin{cases}
    0.5(v)_{+}^{2} &\quad\text{for }h\in\cI_{+},
    \\
    0.5(v)^{2} &\quad\text{for }h\in\cI_{0},
    \\
    0.5(-v)_{+}^{2} &\quad\text{for }h\in\cI_{-},    
  \end{cases}
\end{tsaligned}
both the sides
in \eqref{eq:ub-of-gst-2nd} can be rewritten as
\begin{tsaligned}
  \text{LHS of \eqref{eq:ub-of-gst-2nd}}
  =
  \sum_{h=1}^{d}\barg_{h}(\Delta_{h}\,;\,v_{h}),
\end{tsaligned}
and
\begin{tsaligned}
  \text{RHS of \eqref{eq:ub-of-gst-2nd}}  
  = \sum_{h=1}^{d}\gstar_{h}(v_{h}+\Delta_{h})
\end{tsaligned}
To show the inequality~\eqref{eq:ub-of-gst-2nd}, it suffices to show that
\begin{tsaligned}
  \forall h=1,\dots,d,
  \forall v,
  \forall \Delta\in\bR,\quad
  \barg_{h}(\Delta;v) \ge \gstar_{h}(v+\Delta). 
\end{tsaligned}
Apparently, $\barg_{h}(\Delta;v) = \gstar_{h}(v+\Delta)$
for $h\in\cI_{0}$. Assume $h\in\cI_{+}$ for a while.
\begin{tsaligned}
  \barg_{h}(\Delta\,;\,v)
  =
  \begin{cases}
    0.5(\Delta+v)^{2} &\quad\text{ for }v\ge 0,
    \\
    0.5\Delta^{2} &\quad\text{ for }v< 0. 
  \end{cases}
\end{tsaligned}
The following three cases must be considered:
\begin{itemize}
\item 
In case of $v\ge 0$, 
\begin{tsaligned}
  \barg_{h}(\Delta\,;\,v)
  =
  0.5(\Delta+v)^{2} \ge \gstar_{h}(v+\Delta). 
\end{tsaligned}
\item 
In case of $v< 0$ and $\Delta<-v$,
\begin{tsaligned}
  \barg_{h}(\Delta\,;\,v)
  =
  0.5\Delta^{2} \ge 0 = \gstar_{h}(v+\Delta). 
\end{tsaligned}
\item 
In case of $v< 0$ and $\Delta\ge-v$, 
\begin{tsaligned}
  &\barg_{h}(\Delta\,;\,v)-\gstar_{h}(v+\Delta)
  =
  0.5\Delta^{2} - 0.5(\Delta+v)^{2}
  \\
  &\qquad= - v\Delta - 0.5v^{2}
  = -0.5 ((\Delta+v)+\Delta)v
  \\
  &\qquad\ge -0.5 v \Delta \ge 0.5v^{2}\ge 0. 
\end{tsaligned}
\end{itemize}
Therefore, we get 
$\barg_{h}(\Delta;v) \ge \gstar_{h}(v+\Delta)$
for $h\in\cI_{+}$.  Finally, we assume
$h\in\cI_{-}$. 
\begin{tsaligned}
  \barg_{h}(\Delta\,;\,v)
  =
  \begin{cases}
    0.5(\Delta+v)^{2} &\quad\text{ for }v\le 0,
    \\
    0.5\Delta^{2} &\quad\text{ for }v> 0. 
  \end{cases}
\end{tsaligned}
We need to analyze the following three cases: 
\begin{itemize}
\item 
In case of $v\le 0$, 
\begin{tsaligned}
  \barg_{h}(\Delta\,;\,v)
  =
  0.5(\Delta+v)^{2} \ge \gstar_{h}(v+\Delta). 
\end{tsaligned}
\item 
In case of $v> 0$ and $\Delta>-v$,
\begin{tsaligned}
  \barg_{h}(\Delta\,;\,v)
  =
  0.5\Delta^{2} \ge 0 = \gstar_{h}(v+\Delta). 
\end{tsaligned}
\item 
In case of $v> 0$ and $\Delta\le-v$, 
\begin{tsaligned}
  &\barg_{h}(\Delta\,;\,v)-\gstar_{h}(v+\Delta)
  =
  0.5\Delta^{2} - 0.5(\Delta+v)^{2}
  \\
  &\qquad= - v\Delta - 0.5v^{2}
  = 0.5 (-(\Delta+v)-\Delta)v
  \\
  &\qquad\ge 0.5 (-\Delta)v \ge 0.5v^{2}\ge 0. 
\end{tsaligned}
\end{itemize}
The above leads to 
$\barg_{h}(\Delta;v) \ge \gstar_{h}(v+\Delta)$
for $h\in\cI_{-}$.
\qed


\subsection{Proof of Theorem~\ref{thm:converg-srsdca}}
  \label{ss:proof-thm:converg-srsdca}
A key observation that leads to the discovery of Theorem~\ref{thm:converg-srsdca}
is the following lemma:
\begin{lemma-waku}\label{lem:nabla-convconj-g-corrects-sign}
  Let $g:\bR^{d}\to\bR\cup\{+\infty\}$ be defined as
  $g(\vw) := \frac{1}{2}\lVert\vw\rVert^{2} + \delta_{\cS}(\vw)$ 
  where $\delta_{\cS}(\cdot)$ is the indicator function
  of the feasible region $\cS$ given in \eqref{eq:fearegion}.
  Namely, $\delta_{\cS}(\vw)=+\infty$ if $\vw\not\in\cS$;
  otherwise $\delta_{\cS}(\vw)=0$. 
  Then,
  with $d$-dimensional vector $\vc$ defined in \eqref{eq:scvec-c-def}, 
  the gradient of its convex conjugate~\citep{rockafellar70convex} is expressed
  as
  $\nabla g^{*}(\bar{\vw}) =
    \bar{\vw}
    + \vc\odot(-\vc\odot\bar{\vw})_{+}$. 
\end{lemma-waku}

See Subsections~\ref{ss:proof-lem:nabla-convconj-g-corrects-sign} 
for proof of  
Lemma~\ref{lem:nabla-convconj-g-corrects-sign}. 

The function $g$ defined in
Lemma~\ref{lem:nabla-convconj-g-corrects-sign}
is $1$-strongly convex.  Then, 
if we view $g$ as a regularization function
in replacement of the square L2-norm regularizer,
the sign-constrained optimization
problem~\eqref{eq:prob-rlm-signcon} can be rewritten as
\begin{tsaligned}
  \text{min }\quad&
  \lambda g(\vw) +
  \frac{1}{n}\Phi(\X^\top\vw)
  \qquad\text{wrt }\quad\vw\in\bR^{d}. 
\end{tsaligned}
This is a class of optimization problems targeted by a variant of SDCA named Prox-SDCA~\citep{Shalev-Shwartz2013a-Accelerated} which maintains the convergence rate of the vanilla SDCA yet the regularization function can be extended to be a 1-strongly convex function. The difference from the vanilla SDCA is that the primal variable is recovered from the gradient of the convex conjugate of $g(\cdot)$ at the end of each iterate. It can be seen that Algorithm~\ref{alg:srsdca} is generated by applying Prox-SDCA to our problem setting with $g$ defined in Lemma~\ref{lem:nabla-convconj-g-corrects-sign}. From this observation, Theorem~\ref{thm:converg-srsdca} is established. 
\qed

\subsection{Proof of Lemma~\ref{lem:nabla-convconj-g-corrects-sign}}
\label{ss:proof-lem:nabla-convconj-g-corrects-sign}
The convex conjugate of $g$ is
\begin{tsaligned}
  g_{*}(\bar{\vw})
  &=
  \max_{\vw\in\bR^{d}}
  \left(
  \left<\bar{\vw},\vw\right> - g(\vw)
  \right)
  \\
  &=
  \max_{\vw\in\bR^{d}}
  \left(
  \left<\bar{\vw},\vw\right> - \frac{1}{2}\lVert\vw\rVert^{2}
  - \delta_{\cS}(\vw)
  \right)
  \\
  &=
  \max_{\vw\in\cS}
  \left(
  \left<\bar{\vw},\vw\right> - \frac{1}{2}\lVert\vw\rVert^{2}
  \right)
  \\
  &=
  \frac{1}{2}\lVert\bar{\vw}\rVert^{2}
  -
  \frac{1}{2}
  \min_{\vw\in\cS}
  \lVert\vw-\bar{\vw}\rVert^{2}. 
\end{tsaligned}
We use Danskin's theorem to get the derivative as: 
\begin{tsaligned}
  \nabla g^{*}(\bar{\vw}) = \Pi_{\cS}(\bar{\vw}). 
\end{tsaligned}
\qed

\end{document}